\documentclass[letterpaper, 10pt, conference]{ieeeconf}
\IEEEoverridecommandlockouts
\usepackage[utf8]{inputenc}

\usepackage{comment}
\usepackage{xspace}
\usepackage{graphicx}
\usepackage{overpic}
\usepackage{svg}
\usepackage{xcolor}
\usepackage{bm}
\usepackage[hang,flushmargin]{footmisc}
\usepackage{adjustbox}

\usepackage{tabularx}
\usepackage{multirow}
\usepackage{booktabs}
\usepackage{colortbl}
\usepackage{float}
\usepackage{placeins}

\usepackage{array}
\newcolumntype{P}[1]{>{\centering\arraybackslash}p{#1}}  %

\usepackage{etoolbox}
\makeatletter
\patchcmd{\@makecaption}
{\scshape}
{}
{}
{}
\newcommand{\onetagright}{\tagsleft@false}
\makeatother

\usepackage{amsmath}
\usepackage{amssymb}
\usepackage{amsthm}
\usepackage{siunitx}

\DeclareMathOperator*{\argmin}{arg\,min}

\newtheoremstyle{main}
{1em}                                                %
{1em}                                                %
{\itshape}                                        %
{0pt}                                                %
{\scshape}                                           %
{\\*}                                                %
{2pt}                                                %
{\thmname{#1}\thmnumber{ #2}: \thmnote{\itshape #3}} %

\usepackage{environ}

\usepackage[linesnumbered,ruled,noend]{algorithm2e}
\newcommand{\removelatexerror}{\let\@latex@error\@gobble}

\usepackage[
	activate   = {true},
	protrusion = false,
	expansion  = true,
	kerning    = true,
	spacing    = true,
	tracking   = false,
	auto       = true,
	selected   = true,
	factor     = 1000,
	stretch    = 10,
	shrink     = 10,
]{microtype}

\usepackage{csquotes}
\usepackage[
	maxbibnames=99,
	maxcitenames=2,
	natbib=true,
	bibstyle=ieee,
    citestyle=numeric-comp,
	backend=biber,
	sorting=none,
	giveninits=true,
	url=false,
	doi=false,
	eprint=false,
	isbn=false,
]{biblatex}

\definecolor{purduegold}{HTML}{C28E0E} %

\makeatletter
\let\NAT@parse\undefined
\makeatother
\usepackage[pdfa,colorlinks,bookmarksopen,bookmarksnumbered,allcolors=purduegold]{hyperref}
\usepackage[english]{babel}

\makeatletter
\newcommand\footnoteref[1]{\protected@xdef\@thefnmark{\ref{#1}}\@footnotemark}
\makeatother

\usepackage[nameinlink,capitalise]{cleveref}
\crefname{line}{line}{lines}
\crefname{figure}{Fig.}{Figs.}
\Crefname{figure}{Fig.}{Figs.}
\crefname{equation}{Eq.}{Eqs.}
\Crefname{equation}{Eq.}{Eqs.}
\crefname{section}{Sec.}{Secs.}
\Crefname{section}{Sec.}{Secs.}
\crefname{definition}{Def.}{Defs.}
\Crefname{definition}{Def.}{Defs.}
\crefname{algorithm}{Alg.}{Algs.}
\Crefname{algorithm}{Alg.}{Algs.}
\crefname{assumption}{Asm.}{Asms.}
\Crefname{assumption}{Asm.}{Asms.}
\crefname{subassumption}{Asm.}{Asms.}
\Crefname{subassumption}{Asm.}{Asms.}
\Crefname{problem}{Problem}{Problems}
\crefname{problem}{Problem}{Problems}

\usepackage{flushend}

\let\labelindent\relax
\usepackage[inline]{enumitem}
\usepackage{mathtools}
\usepackage[normalem]{ulem}

\graphicspath{ {./images/} }

\newcommand{\norm}[1]{\left\lVert#1\right\rVert}

\title{\fontsize{17pt}{24pt}\selectfont \bf Physics-Grounded Differentiable Simulation for Soft Growing Robots}
\author{Lucas Chen\authorrefmark{1}, Yitian Gao\authorrefmark{1}, Sicheng Wang, Francesco Fuentes, Laura H. Blumenschein, and Zachary Kingston%
\thanks{
XC, YT, and ZK are with the \href{https://commalab.org/}{CoMMA Lab}, Department of Computer Science, Purdue University.
FF, SW, and LHB are with the \href{https://lhblumen.wixsite.com/website-1}{RAAD Lab}, School of Mechanical Engineering, Purdue University.
Email: \texttt{\{chen4007, gao634, wang5239, ffuente, lhblumen, zkingston\}@purdue.edu}.
}%
\thanks{\authorrefmark{1}Equal Contribution.}
}

\begin{document}
\maketitle
\thispagestyle{empty}
\pagestyle{empty}

\begin{abstract}
Soft-growing robots (i.e., vine robots) are a promising class of soft robots that allow for navigation and growth in tightly confined environments.
However, these robots remain challenging to model and control due to the complex interplay of the inflated structure and inextensible materials, which leads to obstacles for autonomous operation and design optimization.
Although there exist simulators for these systems that have achieved qualitative and quantitative success in matching high-level behavior, they still often fail to capture realistic vine robot shapes using simplified parameter models and have difficulties in high-throughput simulation necessary for planning and parameter optimization.
We propose a differentiable simulator for these systems, enabling the use of the simulator ``in-the-loop'' of gradient-based optimization approaches to address the issues listed above. With the more complex parameter fitting made possible by this approach, we experimentally validate and integrate a closed-form nonlinear stiffness model for thin-walled inflated tubes based on a first-principles approach to local material wrinkling. 
Our simulator also takes advantage of data-parallel operations by leveraging existing differentiable computation frameworks, allowing multiple simultaneous rollouts.
We demonstrate the feasibility of using a physics-grounded nonlinear stiffness model within our simulator, and how it can be an effective tool in sim-to-real transfer. We provide our implementation open source.
\end{abstract}

\section{Introduction}
\label{sec:intro}

Soft and compliant robots naturally conform and deform around objects, which can provide advantages over traditional rigid systems for manipulation~\cite{shintake2018soft} and the traversal of unstructured environments~\cite{calisti2017fundamentals}.
The class of soft robots that grow outward from their tip with pneumatically driven eversion (i.e., vine robots)~\cite{hawkes2017soft} is particularly interesting for applications that require motion through tight and constrained environments~\cite{coad2019vine}, such as search and rescue, non-destructive inspection, and medical biopsy.
However, due to the nonlinearity of the combined structural and material compliance and the frictional interaction with the environment, vine robots are difficult to model without imposing artificial restrictions on the kinematics, making their motion difficult to plan or forecast~\cite{della2023model} and limiting their application in many scenarios.

Contemporarily, advances in physics simulators have brought tremendous advantages to robotics~\cite{collins2021review,makoviychuk2021isaac}, demonstrating the ability to accurately model complex environment interactions and collisions~\cite{pfrommer2021contactnets, tracy2023differentiable}, soft bodies~\cite{du2021underwater}, and more.
Moreover, \emph{differentiable simulation}~\cite{hu2019difftaichi,hu2019chainqueen,le2021differentiable}, in particular efficient parallel differentiable simulation~\cite{ferigo_phd_thesis_2022,jatavallabhula2021gradsim}, has become a key enabling technology for many advances in reinforcement learning~\cite{xu2022accelerated}, parameter fitting for addressing sim-to-real gaps~\cite{heiden2021neuralsim,le2021differentiable}, and more.
These simulators allow for the computation of gradients throughout an entire simulation with respect to a given loss function and simulator parameters, allowing for direct integration into gradient-based optimization frameworks.
Previous work has addressed simulation of growing vine robots using expensive and accurate finite element (FE) methods~\cite{du2023finite,wu2023towards}, video game engines~\cite{li2021bioinspired}, and physically-informed analytical~\cite{blumenschein2021geometric,selvaggio2020obstacle,greer2020robust,fuentes2023mapping} or dynamic~\cite{jitosho2021dynamics} models.
However, while differentiable (e.g.,~\cite{jitosho2021dynamics}), these previous approaches were limited in their predictive capabilities, at least in part due to necessary assumptions in the stiffness model to make the simulations functional. Models of growing robot bending stiffness have assumed either constant moments \cite{selvaggio2020obstacle,greer2020robust} based on empirical results of cantilevered inflated beams \cite{comer1963deflections} or linear stiffness \cite{jitosho2021dynamics}, both of which are decent approximations for some parts of the force, deflection, and pressure space, but neither of which can fully predict a bending behavior \cite{wang2024anisotropic}. %

\begin{figure}
    \centering
    \includeinkscape[width=0.99\linewidth]{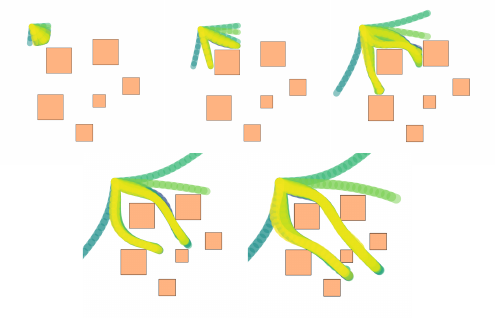}
    \caption{
    Snapshots from a simultaneous rollout of 64 independent vine robots with randomly sampled launch angles in a cluttered environment.
    Each independent trial is uniquely colored.
    Although the rollouts are overlayed, they do not interact with each other in simulation.
    Our simulator is able to efficiently batch operations to compute the dynamics of multiple robots in parallel, as well as retrieve gradients through the simulation.}
    \label{fig:batchsim}
    \vspace{-1.5em}
\end{figure}

We present an end-to-end physics-grounded differentiable simulator for soft-growing vine robots that enables efficient parallel computation and allows the development and integration of a new closed form and experimentally validated inflated beam bending model.
Our simulator is able to accurately reproduce navigation in cluttered environments for these robots, as well as fit the model parameters to predict their local material buckling and bending behaviors.
Compared to other simulators which can provide gradients, we also leverage advances in differentiable optimization~\cite{agrawal2019differentiable}---in particular differentiable quadratic programming~\cite{odonoghue:21} to maintain physical feasibility under constraints for multi-step dynamic simulation.
We demonstrate our simulator's computational efficiency, how our simulator can accurately reproduce trajectories from real systems, and how the gradients from the simulator can be used to optimize physical parameters.
We provide our implementation open source\footnote{\label{fn:url}\url{https://github.com/CoMMALab/DiffVineSimPy}} built on top of PyTorch~\cite{paszke2019pytorch}, enabling integration with existing machine learning frameworks.

\section{Related Work}
\label{sec:related}

\subsection{Vine Robot Modeling}
Vine robots are a class of soft robot that can move by extending in length through pressure-driven eversion, allowing them to extend without sliding against the environment and self-support through the previously grown body~\cite{hawkes2017soft}. 
These robots take a variety of forms, but in general are made from thin, inextensible materials like plastic sheets or air-tight fabrics, which are formed into inflatable tubes~\cite{blumenschein2020design}. 
Due to their material and structure, traditional soft robot modeling approaches that work well for elastomeric materials, such as finite element (FE) modeling, have been challenging to incorporate for vine robots. 
While some FE simulators have been built~\cite{du2023finite,wu2023towards}, they still lack environment interaction and have limited bending of the structure. 

Models which more accurately match the behavior of vine robots have generally been heuristic, based on observation of the robot behaviors, with growth models being a primary focus~\cite{blumenschein2017modeling,coad2020retraction,haggerty2019characterizing}.
Models for bending of the structure and interaction with the environment have evolved from combinations of two assumptions: analytical models of inflated beam buckling~\cite{comer1963deflections} and constant curvature assumptions common to continuum robots.
These have led to models of buckling during environment contact~\cite{haggerty2019characterizing,greer2020robust,fuentes2023mapping}, during retraction~\cite{coad2020retraction} or during unsuspended growth~\cite{mcfarland2023collapse} and quasi-static models of steering with different actuation~\cite{greer2019soft,blumenschein2021geometric}.
These heuristic models have been used to make initial simulations for motion planning \cite{greer2020robust, selvaggio2020obstacle} or environment mapping \cite{fuentes2023mapping}, but they either limit the environment contact~\cite{selvaggio2020obstacle} or the active control of shape~\cite{greer2020robust, fuentes2023mapping}.
To achieve better simulation and planning, we must move past quasi-static assumptions and simplified heuristic models, and towards more realistic nonlinear analytical models of vine robot deflection under force \cite{wang2024anisotropic}.

\subsection{Soft and Vine Robot Simulation}

As discussed previously, analytical characterizations of vine robots can provide powerful predictive frameworks for limited scenarios---understanding vine robot behavior in more complex physical scenarios requires the use of simulation, and the use of accurate simulation can replace and improve on time-intensive and expensive development of robots~\cite{mengaldo2022concise}.
However, in general, simulating soft robots is difficult due to the removal of typical rigidity assumptions present in a physics simulation.
Accurate solutions use FE models built on frameworks like SOFA~\cite{faure2012sofa}, and have been used to model vine robots to great success~\cite{li2021bioinspired,du2023finite,vartholomeos2024lumped,wu2023towards}.
However, these are computationally expensive and ill-suited for real-time applications such as planning and control.
Simplified rigid body approximate models have also been used to model vine robots, either using minimal coordinates (virtual joint angles) (e.g.,~\cite{el2018development}) or maximal coordinates (poses of frames in a global reference with implicit constraints) (e.g.,~\cite{jitosho2021dynamics}).
For reasons of efficiency and robustness, our simulator uses a maximal coordinate representation, based \mbox{off~\citet{jitosho2021dynamics}}.

\subsection{Differentiable Simulation}
Differentiable simulation is a powerful emerging framework that has found great success in robotics~\cite{collins2021review} due to the ongoing explosion in automatic differentiation~\cite{baydin2018automatic}, efficient parallel processing libraries for deep learning (e.g., PyTorch~\cite{paszke2019pytorch}, JAX~\cite{bradbury2018jax}) enabling massively parallel simulation~\cite{freeman2021brax}, and differentiable computation~\cite{hu2019difftaichi}.
The key advantage provided of differentiable simulation is the ability to ``crack open'' the black box of the simulation, providing gradients directly from a desired loss function back to the inputs or parameters of the simulation, potentially over multiple time steps.
This enables the use of far more efficient gradient-based optimization approaches for solving inverse problems~\cite{coevoet2017optimization}, advantages in system identification~\cite{lee2023robot,le2021differentiable}, reinforcement learning~\cite{xu2022accelerated,suh2022differentiable}, planning to exploit contact and long-horizon physical effects~\cite{toussaint2018differentiable}, and embedding neural or learned functions inside the physical simulation itself~\cite{heiden2021neuralsim}.
Advances in differentiable optimization~\cite{odonoghue:21,agrawal2019differentiating,amos2017optnet, agrawal2019differentiable} have also enabled better physical feasibility in differentiable simulation (e.g.,~\cite{howell2022dojo}).
Our simulator also solves an internal quadratic program to find feasible velocities.

\subsection{Differentiable Soft Robot Simulation}

Differentiable simulation has seen incredible success in soft robotics, particularly in deformable objects and robots~\cite{hu2019chainqueen}, either through physical equations or through neural surrogate models~\cite{bern2020soft}.
Due to the high compliance and low stiffness of the system, gradients returned from the simulator are more informative and can be used in a variety of contexts, e.g., automatic controller~\cite{du2021underwater} and robot design~\cite{ma2021diffaqua} for underwater swimmers, optimizing robot designs~\cite{spielberg2019learning}, and matching physics directly from video through differentiable rendering~\cite{jatavallabhula2021gradsim}.

We note that~\citet{jitosho2021dynamics} used the automatic differentiation capabilities of Julia's \texttt{ForwardDiff.jl} library~\cite{RevelsLubinPapamarkou2016} to perform gradient-based optimization of fitting the parameters of the vine robot simulation to video.
However, this approach used a single-step fixed time-step model of the robot's dynamics, unlike our approach---a general-purpose differentiable simulator for vine robots.

\section{Differentiable Vine Robot Simulation}
\label{sec:method}

\subsection{Revisiting Bending Stiffness Assumptions}
\label{sec:bending}

\begin{figure}
    \centering
    \includegraphics[width=\linewidth]{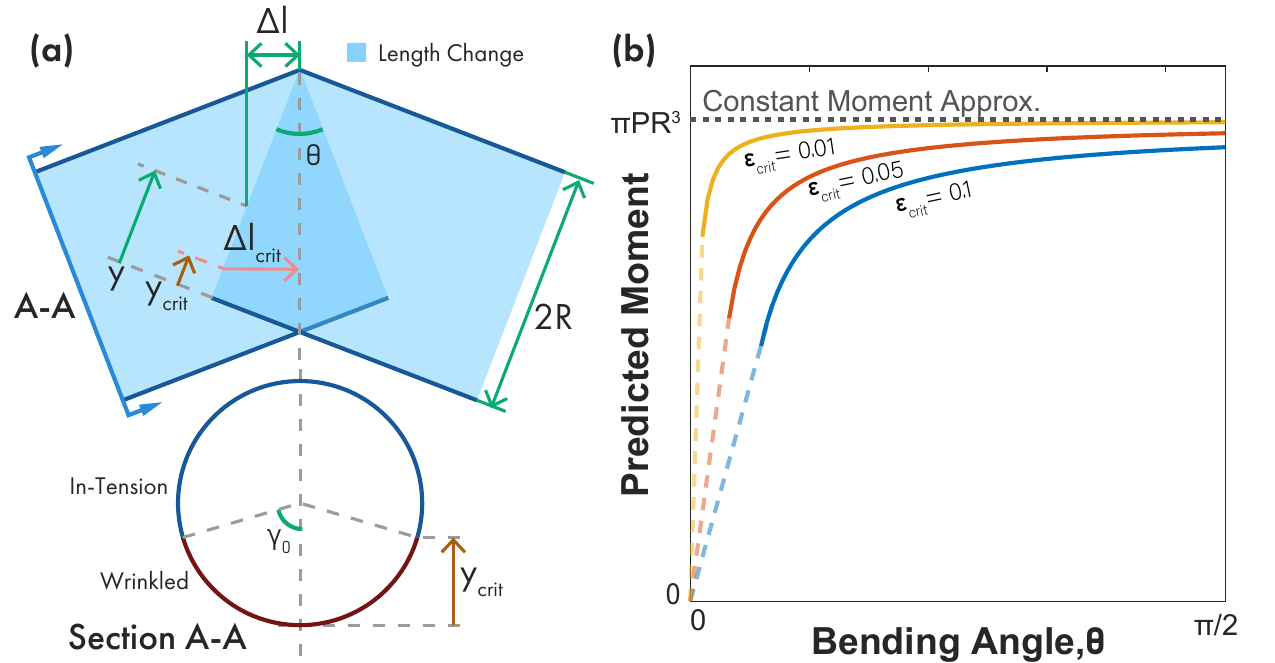}
    \caption{(a):The geometry at the joint of an inflated beam with variables in the joint model. Section A-A further shows the division of the wrinkled and tensioned regions. (b): Variation of the moment at various values of $\epsilon_{crit}$, as compared with the fully wrinkled moment, $\pi PR^3$.}
    \label{fig:bending_geom}
    \vspace{-1.5em}
\end{figure}

Modeling the stiffness of the vine robot is central to capturing its behavior. Previous dynamic simulations~\cite{jitosho2021dynamics} have modeled vine robot as a series of rigid bodies connected by damped linear torsional springs. However, unlike the uniform, isotropic backbone found in many continuum robots, the response of the inflated body of vine robots to external forces deviates significantly from linear-elastic behavior, and is dominated instead by local loss of tension (wrinkling) at the surface: the wrinkling-based model by~\citet{comer1963deflections} states that the moment induced in bending an inflated beam is proportional to the height of wrinkling and approaches some constant final value. In several prior works this model was simplified as a constant bending moment~\cite{haggerty2019characterizing, mcfarland2023collapse}, but this leads to significant overestimation of the stiffness, and therefore interaction forces, for small bending angles.

Obtaining a more refined model of the restoring moment for use in vine robot simulation requires relating the development of wrinkling to the joint bending angle, since the first principles model developed in~\cite{comer1963deflections} defines the moment as a function of the amount of wrinkling. The bending moment can be calculated as~\cite{comer1963deflections}
\begin{equation}
    M= \pi PR^3\frac{\sin2\gamma_0+2\pi-2\gamma_0}{4[\sin\gamma_0+(\pi-\gamma_0)\cos\gamma_0]},
    \label{eqn:comerBeam}
\end{equation}
where $\gamma_0$ is the angle corresponding to the wrinkled surface, as shown in Section A-A of~\cref{fig:bending_geom}(a).
We bridge the gap between the wrinkling angle $\gamma_0$ and the bending angle $\theta$ by introducing a non-dimensional wrinkling criterion, $\epsilon_{crit}=\Delta l_{crit} /{2R}$, which gives the ratio between a length change threshold $\Delta l_{crit}$ at which wrinkling occurs and the radius of the inflated beam. As shown in~\cref{fig:bending_geom}(a), the overlapping region formed by the two initially non-intersecting volumes suggests a change of length $\Delta l$ at some height $y$ on the surface. The critical height $y_{crit}$ where $\Delta l=\Delta l_{crit}$ relates readily to $\gamma_0$ by geometry. In terms of the wrinkling criterion,
\begin{equation}
    \gamma_0 = \cos^{-1}(2\frac{\epsilon_{crit}}{\sin(\theta/2)}-1).
    \label{eqn:gam}
\end{equation}
Substituting~\cref{eqn:gam} into~\cref{eqn:comerBeam}, we obtain a modified, bending angle-dependent moment, the value of which is shown in comparison to the full bending moment ($\pi PR^3$) at selected values of $\epsilon_{crit}$ in~\cref{fig:bending_geom}(b). A smaller value of $\epsilon_{crit}$ results in a more rapid increase in bending moment and a final value closer to the full bending moment, implying less resistance to wrinkling formation and hence a more rapid development with increasing bending angle. It can be further noted that \cref{eqn:gam} is only valid for $\theta \geq 2\sin^{-1}(\epsilon_{crit})$, i.e., not valid when the bending angle is not large enough to induce wrinkling. For model completeness, we assume the beam exhibits linear-elastic behaviors before wrinkling.

\begin{figure}
    \centering
    \includegraphics[width=\linewidth]{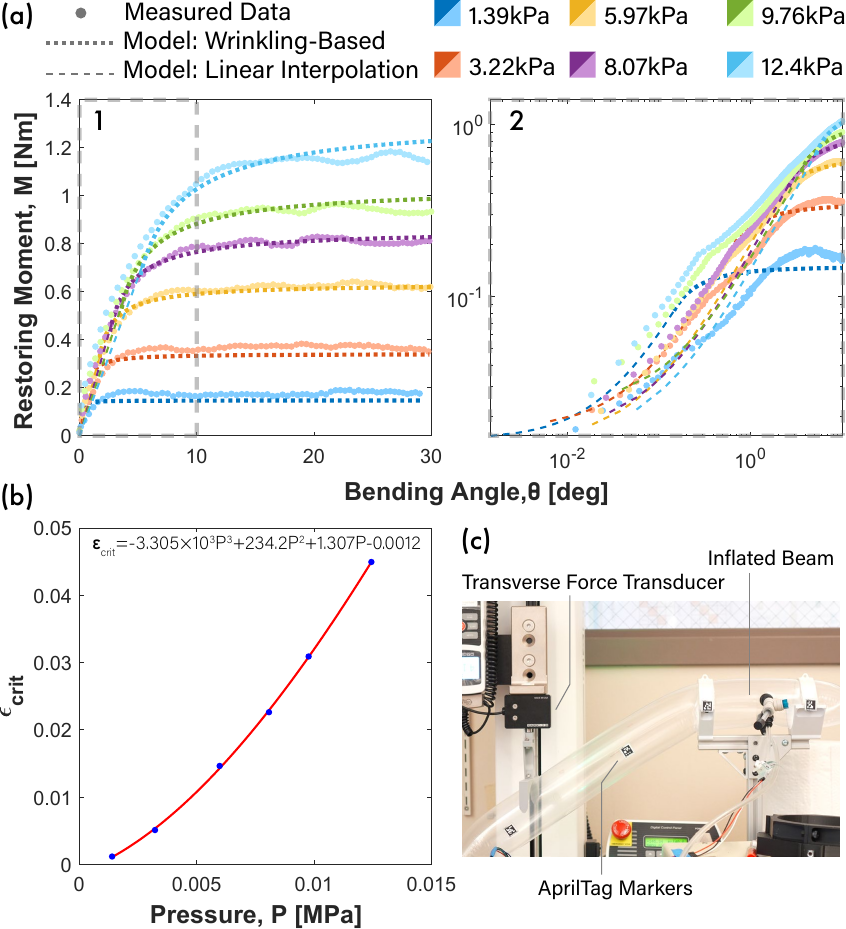}
    \caption{(a): Experimentally measured bending moment of an inflated beam segment and the proposed model prediction. (a).1 shows the full range of bending angles (with the measured data down-sampled for legibility), and (a).2 shows a magnified view of the initial increase of the restoring moment in log scale. (b): Wrinkling criterion $\epsilon$ obtained for various pressures and fitted third-degree polynomial. (c): Experimental setup.}
    \label{fig:bending_exp}
    \vspace{-1.5em}
\end{figure}

The value of $\epsilon_{crit}$ used in the simulation is obtained experimentally for the LDPE tubing used to make the vine robot in~\cref{sec:real}. The setup shown in~\cref{fig:bending_exp}(c) allows accurate measurement of applied force and resulting bending angle. The LDPE tubing (67mm diameter, 2mil, ULine) is fixed at a known distance from a force transducer (MR03-5, Mark-10 Inc.) mounted on a traveling stand (EM-303, Mark-10 Inc.). The end-effector of the traveling stand is pushed down onto the inflated tubing at a rate of 150mm/s. The pressure in the tubing was maintained with a pressure regulator (QB3, Proportion Air). At each input pressure, $\epsilon_{crit}$ is found via a least-square fitting to the data, 
\begin{equation*}
    \argmin_{\epsilon_{crit}} (\mathbf{M}_{measured}(\theta)-\mathbf{M}_{predict}(\theta,\epsilon_{crit}))^2
\end{equation*}
\cref{fig:bending_exp}(a) shows the experimental results and the best-fit model. The wrinkling criterion-based prediction captures the evolution of the bending moment well, especially at relatively high input pressures. For small angles, the linear elasticity assumption under and over-predicts the bending moment at higher and lower pressures respectively. This may be remedied by considering small deformation beam mechanics in the future. For use in the simulation, $\epsilon_{crit}$ was precomputed at various pressures and a heuristic third-degree polynomial was fit, as shown in~\cref{fig:bending_exp}(b).

\subsection{Simulation Formulation}
\label{sec:sim}

\begin{figure}[t]
    \centering
    \includeinkscape[width=0.8\linewidth]{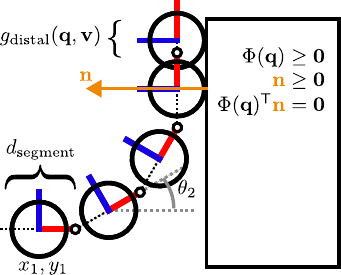}
    \caption{
    Illustration of the virtual rigid body model of the vine robot.
    A series of frames parameterized by $(x, y, \theta)$, each with a collision sphere, are connected by pin joints a fixed distance $d_\text{segment}$ apart.
    The final link is the only exception: it grows with respect to the $g_\text{distal}$ function.
    Contact complementarity conditions enforce collisions with the environment.}
    \label{fig:model}
\end{figure}

We build upon the prior work of~\citet{jitosho2021dynamics} on dynamic simulation of vine robots as a collection of rigid bodies.
This previous work makes a number of choices in the design of their simulator that we also use, such as an impulse-velocity dynamics formulation, a maximal coordinate representation of poses of the virtual links of the robot, and a Lagrange multiplier formulation.

We reproduce parts of~\citet{jitosho2021dynamics} here for convenience. For additional details, please see the original work.
The Lagrange multiplier formulation uses a configuration $\mathbf{q}$ and velocities $\mathbf{v}$ in maximal coordinates, composed of frame poses $(x, y, \theta)$ for each virtual joint, which describes the pose of the center of mass for the virtual link---each link has a collision sphere with diameter equal to $d_\text{segment}$.
Constraints on the system are described by sets of constraint equations $c(\mathbf{q}) = \mathbf{0}$, which represents constraint errors on $\mathbf{q}$ and thus is satisfied when equal to zero.
Violations in constraints result in an impulse vector $\mathbf{\lambda}$, which is mapped into maximal coordinates by the Jacobian of the constraint equation, $J(\mathbf{q}) = \frac{\partial c}{\partial \mathbf{q}}(\mathbf{q})$.
There are also contact complementarity conditions~\cite[Eq. 5]{jitosho2021dynamics} which prevent penetration between the contact forces on the vine and the obstacles. Specifically, $\Phi(\mathbf{q})$ computes a vector of distances between the contact points on the vine to the closest surface in the environment, and $\mathbf{n}$ is a vector of the normal forces exerted from the environment on the contact points. Together, these constraints ensure that penetrating configurations are invalid, the environment only exerts repulsive forces on the vine, and such forces can only happen when the contact points are touching a surface.
\begin{equation*}
\Phi(\mathbf{q}) \geq \mathbf{0}, \qquad
\mathbf{n} \geq \mathbf{0}, \qquad
\Phi(\mathbf{q})^\mathsf{T}\mathbf{n} = 0,
\end{equation*}
where $L(\mathbf{q}) = \frac{\partial \Phi}{\partial \mathbf{q}}(\mathbf{q})$.

In contrast to~\citet{jitosho2021dynamics}, our simulator uses a variable number of virtual links to better capture the everting growth of the vine robots.
We provide a maximum number of links that determines the internal tensor sizes used, although this can be resized dynamically.
Moreover, the use of a variable length vector provides several advantages.
First, it allows our model to avoid modeling prismatic elements between virtual joints, as we assume that each virtual link is equidistant from the prior by a fixed distance $d_\text{segment}$, minus the final element that is currently growing.
This halves the number of links that need to be modeled and reduces the total number of constraints imposed on the system.
Second, constant length simplifies the computation of torques in our bending model, since the moment of inertia is constant.
We depict our formulation with an illustration in~\cref{fig:model}.

Thus, the robot is represented by a series of frames, each of which has a point mass and a collision sphere centered at the origin, enforcing the following revolute constraint per pair of frames~\cite[Eq. 2]{jitosho2021dynamics}:
\begin{equation*}
\begin{bmatrix}
x_k + d\cos(\theta_k) - x_{k+1} - d\cos(\theta_{k+1}) \\
y_k + d\sin(\theta_k) - y_{k+1} - d\sin(\theta_{k+1})
\end{bmatrix} = \mathbf{0}
\end{equation*}
The only link that does not have a fixed distance to its neighbor is the one at distal tip, which grows smoothly according to the velocity \( g_{distal}\):
\begin{equation*}
\resizebox{\columnwidth}{!}{
$g_\text{distal}(\mathbf{q}, \mathbf{v}) = \dfrac{(x_n - x_{n - 1})(\dot{x}_n - \dot{x}_{n - 1}) + (y_n - y_{n - 1})(\dot{y}_n - \dot{y}_{n - 1})}{\sqrt{ (x_n - x_{n - 1})^2 + (y_n - y_{n - 1})^2 }},$}
\end{equation*}
where $n$ is the index of the distal link. This link grows until it reaches a maximum length, then an additional link is added to the vine, with the corresponding dynamics multipliers, $\mathbf{w}$, $G(\mathbf{q}) = \frac{\partial g}{\partial \mathbf{q}}(\mathbf{q})$.

Critically, we also use different models of stiffness than the prior damped linear model:
\begin{equation}
F(\mathbf{q}, \mathbf{v}) = R \tau(q, v) = R(-K(q, \cdot) - C \dot{\theta}(v)),
\end{equation}
where $K(\cdot)$ is a stiffness function (in~\citet[Eq. 7]{jitosho2021dynamics}, this is a linear matrix applied to joint angles, in this work, as described in~\cref{sec:bending}), $\theta(q)$ are pin joint angles given a configuration, $C$ is a damping matrix, $\dot{\theta}(q)$ are pin joint velocities, and $R$ is a transformation of torques into maximal coordinates.

The dynamics of the system are thus the following subject to constraints:
\begin{equation*}
\label{eq:dyn}
    M (\mathbf{v}_{k+1} - \mathbf{v}_k) = J(\mathbf{q})^\mathsf{T} \mathbf{\lambda} + L(\mathbf{q})^\mathsf{T} \mathbf{n} + G(\mathbf{q})^\mathsf{T} \mathbf{w} + F(\mathbf{q}, \mathbf{v}) \Delta t,
\end{equation*}
where $M$ is a mass matrix, $\Delta v = v_{k+1} - v_k$ is change in velocity, $J(q)^\mathsf{T} \mathbf{\lambda}$ is the mapping of constraint impulses into maximal coordinates, $F$ is other forces, and the $\Delta t$ is a time step.
We follow the linearization taken in~\citet{jitosho2021dynamics}, resulting in the solution of the following quadratic program giving the velocity to propagate the system to the next time step:
\begin{align*}
    \min_{\mathbf{v}_{k+1}} \qquad & \dfrac{1}{2} \mathbf{v}_{k+1}^\textsf{T} M \mathbf{v}_{k+1} - \mathbf{v}_{k+1}^\textsf{T} (M \mathbf{v}_{k} + F(\mathbf{q}_k, \mathbf{v}_k) \Delta t) \\ 
    \text{subject to} \qquad & c(\mathbf{q}_k) + J(\mathbf{q}_k)\mathbf{v}_{k+1} \Delta t = \mathbf{0} \\
    & \Phi(\mathbf{q}_k) + L(\mathbf{q}_k)\mathbf{v}_{k+1} \Delta t \geq \mathbf{0} \\
    & g_\text{distal}(\mathbf{q}_k, \mathbf{v}_k) + G(\mathbf{q}_k)\mathbf{v}_{k+1} \Delta t - \mathbf{u}_k = \mathbf{0} \\
\end{align*}

\subsubsection{Differentiable Quadratic Programming}
As mentioned in the prior section, feasible velocities for the next time step are solved by finding the solution to a linearized feasibility problem, which can be modeled as a quadratic program (QP) with linear equality and inequality constraints.
We take advantage of advances in \emph{differentiable optimization} to be able to pass gradients through not only the physics model, but also through general forms of QPs:
\begin{align*}
    z_{i+1} =& \argmin_z & \frac{1}{2} z^T Q(z_i) z + q(z_i)^T z && \qquad \\
    & \text{subject to} & A(z_i)z = b(z_i) &&\\
    & & G(z_i)z \leq h(z_i) &&
\end{align*}
Where $z$ is the optimization variable, and $Q(z_i)$, $q(z_i)$, $A(z_i)$, $b(z_i)$, $G(z_i)$, and $h(z_i)$ are parameters of the optimization problem.

Concretely, we use the Splitting Conic Solver (SCS)~\cite{odonoghue:21} through CVXPyLayers~\cite{agrawal2019differentiable}.
Note that the Anderson acceleration~\cite{aa2020} for SCS has to be appropriately tuned for solving the linear feasibility problem: high values of lookback\footnote{\url{https://www.cvxgrp.org/scs/algorithm/acceleration.html}} are required when stiffness of the system is high.
We note that while CVXPyLayers is effective and competitive with other QP solvers, especially in the batched context, it is not fully GPU-accelerated, thus bounding the performance of our approach.
It is of interest in future work to investigate, e.g., neural surrogates of the QP as in~\citet{heiden2021neuralsim} to accelerate performance.

\subsubsection{Implementation Details}
We have implemented our simulator in PyTorch~\cite{paszke2019pytorch}, a popular framework for machine learning and differentiable computing.
Taking advantage of efficient data-parallel operations through automatic transformations through PyTorch.
We note the existence of other frameworks that are capable of these transformations, e.g., Difftaichi~\cite{hu2019difftaichi}, JAX~\cite{bradbury2018jax}---we chose PyTorch due to the ubiquity of support, as well as with the future goal in mind of using our simulator ``inside the loop'' for training neural policies and behaviors.

\section{Simulation and Fitting Results}
\label{sec:experiments} 
We evaluate our simulator on a computer with a Intel\textsuperscript{\textregistered} Core\textsuperscript{\texttrademark} i9-14900K with 64GB of RAM.
As CVXPyLayers is not GPU-accelerated, we use the CPU for all experiments.

\subsection{Simulator Performance}

\begin{figure*}
    \centering
    \includeinkscape[width=.99\linewidth]{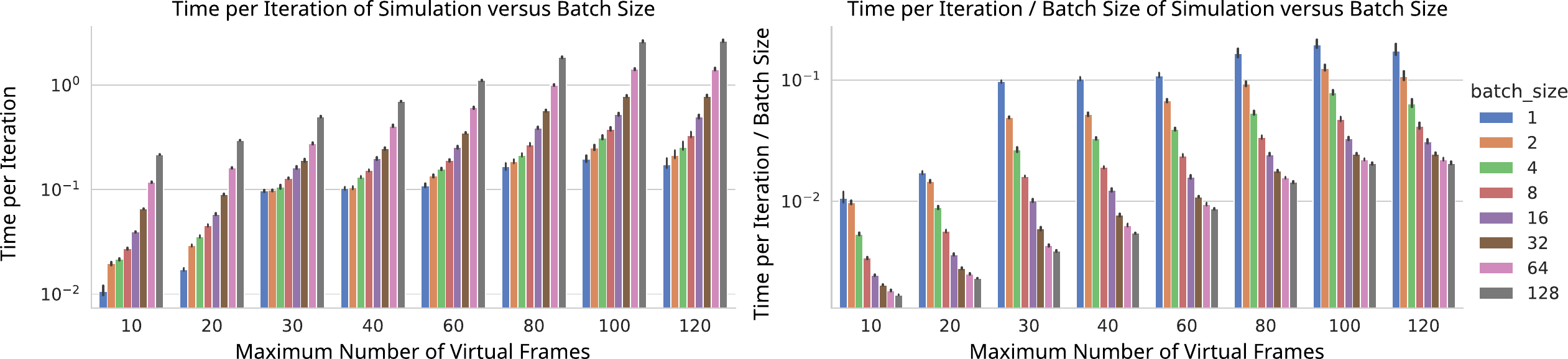}
    \caption{Performance of our simulator averaged over 5 rollouts with batches of randomly sampled launch angles.
    The environment used is shown in~\cref{fig:batchsim}, which shows an example rollout with a batch size of 64 and 40 maximum virtual links.
    On the left, time per iteration of the simulator is given for different numbers of maximum virtual links and batch sizes.
    On the right, time per iteration divided by the batch size (i.e., how efficient is the parallelism) is given.
    }
    \label{fig:performance}
\end{figure*}

We evaluate the time per iteration of the solver solving for the next time step, with varying numbers of maximum virtual links and batch size---results are shown in~\cref{fig:performance}, and an example rollout is shown in~\cref{fig:batchsim}. Rollouts were collected from 5 random trials with random launch angles for every robot in the batch, with 100 time steps each.
Results show our simulator scales well with increasing number of frames, and that increasing the batch size improves total throughput of the simulator in simulating more vine robots faster than smaller batch sizes.
We note that our approach is currently CPU bound---there exist quadratic solvers such as~\cite{amos2017optnet} that are GPU-accelerated and support backpropagation, and it is of interest to investigate potential solutions to move fully to the GPU.

\subsection{Real Robot Video Data}
\label{sec:real}

\begin{figure*}
    \centering
    \includeinkscape[width=.9\linewidth]{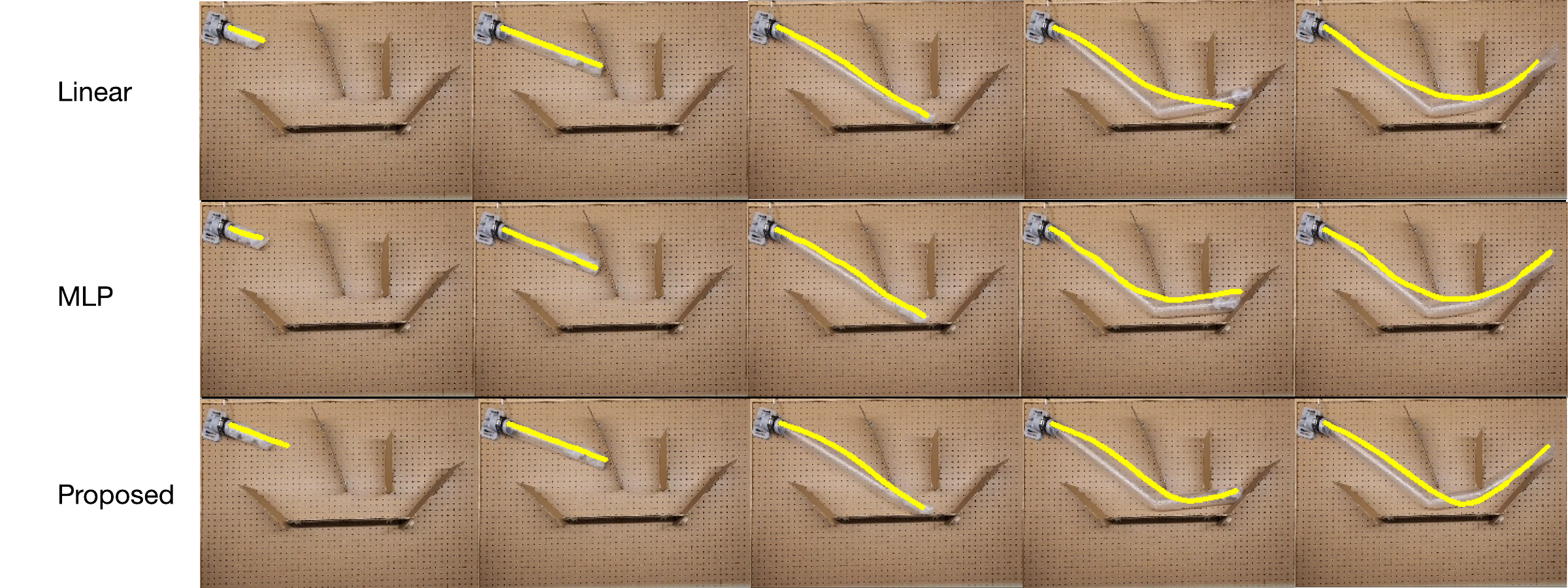}
    \caption{
    Overlay of simulator rollouts in an testing environment not used for parameter fitting.
    Three models were tested, from top to bottom: linear stiffness, a multi-layer perceptron, and the proposed stiffness model.
    Parameters were learned through the optimization routine described in~\cref{sec:fitting}.
    The proposed stiffness model captures the non-constant curvature bending of vine robots when contacting obstacles~\cite{greer2020robust} more accurately than other models.
    }
    \label{fig:realfit}
\end{figure*}

This work was motivated in part by being able to create a generalizable simulator model of vine robots that can be used to predict robot behavior in novel environments.
We collected 10 physical demonstrations of a real vine robot at varying launch positions in 4 environments, with the goal of capturing different behaviors, even within the same environment, due to launch positioning. The robots were created with LDPE tubes with a thickness of 0.002~in and given a constant pressure throughout its deployment.

Raw videos were processed using a hybrid human- and machine-annotated pipeline.
The starting reference frame, environment's bounding box, obstacles, and the vine robots' base are human-annotated for each trial.
To extract reference virtual joint poses in maximal coordinates, we process subsequent frames of the video using OpenCV~\cite{opencv_library}.
A pixel difference is taken each frame with the reference frame, isolating the motion of the robot, which is the processed with blurring and morphological closure to handle camera noise.
The vine's shape is then skeletonized into a curve and divided into $n$ points based on the desired link length, which are then transformed relative to the robot's base, thus providing a dataset of ground truth positions that can be fit and evaluated against.

\subsection{Fitting Parameters}
\label{sec:fitting}

\begin{figure}
    \centering
    \includegraphics[width=\linewidth]{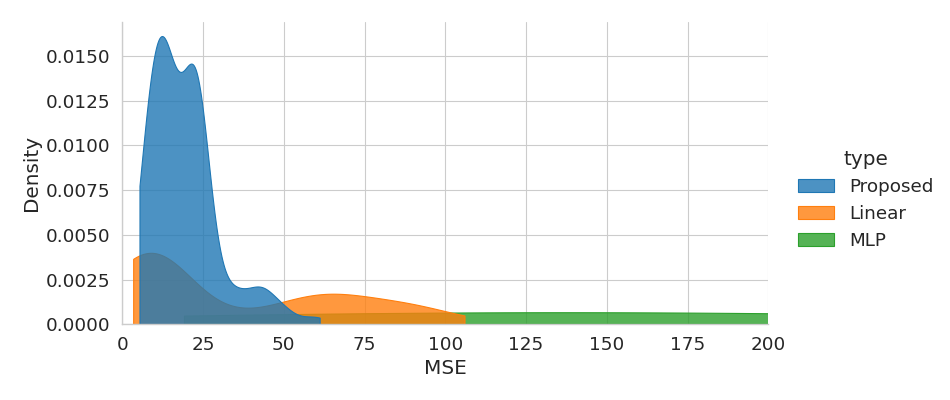}
    \caption{Kernel density estimate of mean-squared error for fitting for linear, multi-layer perceptron, and proposed stiffness models over the course of a rollout.
    The proposed method is more consistent than either approach.}
    \label{fig:mse}
\end{figure}

As our simulator is differentiable, it is easy to plug into any optimization pipeline with customized loss functions.
We use AdamW~\cite{loshchilov2017fixing} as the optimizer of choice for all our experiments, although any solver would work.
Within our simulator, we evaluate parameter fitting with three internal models of stiffness: the linear model from~\citet{jitosho2021dynamics}, a multi-layer perceptron (a two-layer network with 10 hidden units and a hyperbolic tangent activation function), and the proposed stiffness model described in~\cref{sec:bending}.

For fitting, we want to optimize the mass $m$, inertia $I$, growth rate $u$, damping $C$, and stiffness parameters. Although we have already empirically measured the stiffness parameter for one robot in~\cref{sec:bending}, we derive this parameter again on a different vine robot using fitting to demonstrate the capabilities of our model.
\begin{equation*}
    \argmin_{m, I, u, K, C, v_i} \sum^n_{i = 1} \norm{\hat{q}_{i+1} - f(q_i, v_i)}^2_2
\end{equation*}
In the above, we have the forward model $f()$, and try to minimize the \emph{positional} squared error against the ground truth.
We fit on real robot data as described in~\cref{sec:real}. %
We evaluate our fit parameters on unseen data, an example rollout of which is shown in~\cref{fig:realfit}, with mean-squared error data given in~\cref{fig:mse}.

\section{Discussion}
\label{sec:discussion}

We have presented an end-to-end differentiable simulation framework for soft growing robots implemented in PyTorch, enabling efficient parallel computation. Using this improved simulation framework, we also proposed and integrated an improved closed-form model of inflated beam bending and buckling which lead to more realistic growing robot shapes prediction.
We demonstrate the effectiveness of our simulator and stiffness model by showing efficient performance in batch computation as well as accurately fitting our stiffness model parameters to predict real robot execution.
Our implementation is available open source\footnoteref{fn:url}.
We hope that our simulator will be a useful tool in development of planning and control techniques for vine robots, such as in shape optimization, differentiable model predictive control~\cite{amos2018differentiable}, and that our non-linear stiffness model can inspire better understanding of hard to predict vine robot obstacle interactions.

In the future, we wish to address limitations in our current simulation framework relating to unrealistic physics by modeling more complex interactions or by integrating neural models within the simulation (e.g., as done in~\cite{heiden2021neuralsim}).
We also wish to add different modes of actuation used in the design and control of vine robots, such as pre-formed welds, pinches, tensioning cables, and more. Within the stiffness model, we hope the expand the model to better predict the effects of surface mounted actuation, as well as better understand the role of material, geometry, and pressure in shaping the model.

We note limitations in our current simulation relating to instability with high stiffness---other differentiable simulation frameworks have noted and addressed this~\cite{suh2022differentiable}.
Contacts are also difficult to handle well, and using more intelligent techniques (e.g.,~\cite{le2023single,antonova2023rethinking}) are of interest, especially moving to more complex contact primitives~\cite{tracy2023differentiable}.

\printbibliography{}

\end{document}